\documentclass[letterpaper, 10 pt, conference]{ieeeconf}  

\IEEEoverridecommandlockouts                              

\usepackage{graphicx}
\usepackage{booktabs}
\usepackage{makecell}
\usepackage{amsmath}
\usepackage{amssymb}
\usepackage{caption}
\usepackage{diagbox}
\usepackage{multirow}
\usepackage[table,dvipsnames]{xcolor}
\usepackage{hyperref}
\hypersetup{
    colorlinks=true,
    linkcolor=blue,
    urlcolor=magenta
    }

\makeatletter
\def\blfootnote{\gdef\@thefnmark{}\@footnotetext}
\makeatother

\newcommand{\argmin}{\operatornamewithlimits{argmin}}

\usepackage{textcomp}
\usepackage{color,colortbl}
\definecolor{beaublue}{rgb}{0.74, 0.83, 0.9}

\title{\LARGE \bf A Hierarchical Spatiotemporal Action Tokenizer for In-Context Imitation Learning in Robotics
}

\author{Fawad Javed Fateh$^\dagger$~~~~~Ali Shah Ali$^\dagger$~~~~~Murad Popattia~~~~~Usman Nizamani~~~~~Andrey Konin\\M. Zeeshan Zia~~~~~Quoc-Huy Tran
   \thanks{All authors are with Retrocausal, Inc., Redmond, WA 98052, USA.\newline
	Website: \url{www.retrocausal.ai}\newline
	Email: {\tt\small \{fawad,alishah,murad,usman,andrey,\newline
	zeeshan,huy\}@retrocausal.ai}}}

\begin{document}

\maketitle
\thispagestyle{empty}
\pagestyle{empty}

\begin{abstract}
We present a novel hierarchical spatiotemporal action tokenizer for in-context imitation learning. We first propose a hierarchical approach, which consists of two successive levels of vector quantization. In particular, the lower level assigns input actions to fine-grained subclusters, while the higher level further maps fine-grained subclusters to clusters. Our hierarchical approach outperforms the non-hierarchical counterpart, while mainly exploiting spatial information by reconstructing input actions. Furthermore, we extend our approach by utilizing both spatial and temporal cues, forming a hierarchical spatiotemporal action tokenizer, namely HiST-AT. Specifically, our hierarchical spatiotemporal approach conducts multi-level clustering, while simultaneously recovering input actions and their associated timestamps. Finally, extensive evaluations on multiple simulation and real robotic manipulation benchmarks show that our approach establishes a new state-of-the-art performance in in-context imitation learning.
\end{abstract}

\section{Introduction}
\label{sec:introduction}

\begin{figure*}[t]
	\centering
		\includegraphics[width=0.85\linewidth, trim = 0mm 20mm 0mm 0mm, clip]{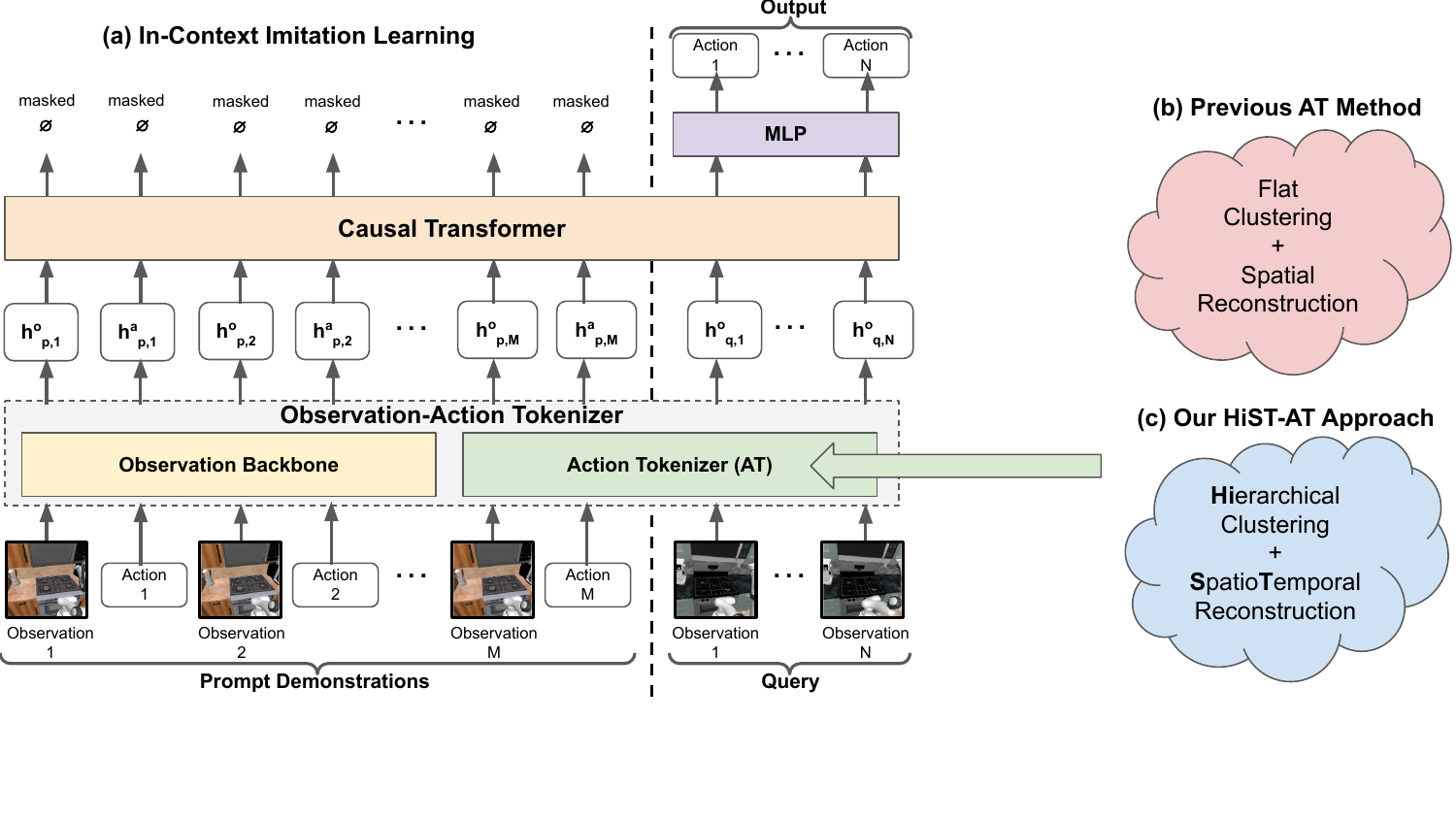}
	\caption{(a) In-context imitation learning (ICIL)~\cite{fu2024context} allows robots to generalize from demonstrations to new tasks without retraining. Action tokenizer (AT) is important to capturing demonstration information effectively. (b) Previous AT methods (e.g., \cite{vuong2025action}) rely on vector quantization, conducting flat clustering and focusing on spatial cues via recovering input actions. (c) We propose a hierarchical spatiotemporal action tokenizer, which performs multi-level clustering and exploits both spatial and temporal cues by jointly reconstructing actions and timestamps, yielding superior performance.}
	\label{fig:teaser}
\end{figure*}

Teaching robots to perform actions from demonstrations has received significant research interest alongside advances in deep learning. A prominent paradigm, imitation learning (IL), aims to learn generalizable robot policies from expert demonstrations~\cite{mandlekar2021matters}. However, as mentioned in~\cite{team2024octo}, IL suffers from limited generalization due to the scarcity of high-quality demonstrations. Recent large-scale efforts~\cite{team2024octo,o2024open,khazatsky2024droid} attempt to alleviate this issue; however, adapting to new tasks often still requires collecting additional task-specific data for fine-tuning. Inspired by the in-context learning capabilities of large language models (LLMs)~\cite{brown2020language,mirchandani2023large,vosylius2023few,kwon2024language,vosylius2025instant}, in-context imitation learning (ICIL)~\cite{di2024keypoint,papagiannis2024r,yin2024context,fu2024context,zhang2025dynamics} has emerged as a promising alternative. ICIL allows robotic policies to perform new tasks from demonstrations provided at inference time, without retraining, enabling flexible and efficient real-world deployment. Fig.~\ref{fig:teaser}(a) shows an example ICIL framework~\cite{fu2024context}.

Despite its advantages, ICIL still struggles to learn contextualized action representations from demonstrations~\cite{park2025incontext}. Effective action representations can lead to notable performance gains in ICIL~\cite{wang2023large}. Several works~\cite{brohan2022rt,shafiullah2022behavior,zhao2023learning,o2024open,huang2024emotion,kim2024openvla,fu2024context,zhang2025dynamics,pertsch2025fast} focus on action tokenizers for discretizing and encoding robot actions, with modeling temporal correlations remaining a key challenge. While positional encoding~\cite{vaswani2017attention} or vector quantization~\cite{yu2024language} can be used to preserve temporal order, they often fail to maintain temporal smoothness in action trajectories~\cite{bharadhwaj2024roboagent}. As discussed in Mysore et al.~\cite{mysore2021regularizing}, temporal smoothness promotes continuity in tokenized actions and reduces noise. Recently, LipVQ-VAE~\cite{vuong2025action} proposes an action tokenizer built on a vector-quantized autoencoder (VQ-VAE)~\cite{van2017neural} and enforces temporal smoothness through Lipschitz regularization. It performs flat clustering of input actions and mainly exploits spatial cues by reconstructing actions (see Fig.~\ref{fig:teaser}(b)).

Motivated by the success of spatiotemporal reconstruction~\cite{kukleva2019unsupervised,vidalmata2021joint} and hierarchical vector quantization~\cite{spurio2025hierarchical,gokay2025skeleton} in temporal action segmentation, we propose a novel hierarchical spatiotemporal action tokenizer (HiST-AT) for in-context imitation learning (see Fig.~\ref{fig:teaser}(c)). First, our approach performs clustering across multiple vector quantization levels, enabling the discovery of short-term sub-action primitives that combine to form long-term coherent actions. Second, it jointly reconstructs input actions along with their timestamps, leveraging both spatial and temporal cues through explicit modeling. As a result, HiST-AT is capable of extracting hierarchical action structures and spatiotemporal dependencies, yielding effective and transferable action representations. Finally, we conduct extensive evaluations on simulation (RoboCasa~\cite{nasiriany2024robocasa} and ManiSkill~\cite{tao2024maniskill3}) and real robotic manipulation datasets, demonstrating superior performance and generalization over prior methods.

In summary, our contributions include:
\begin{itemize}
    \item We first develop a hierarchical action tokenizer for in-context imitation learning based on hierarchical vector quantization. Our hierarchical approach outperforms the non-hierarchical baseline, while focusing on spatial cues via reconstructing input actions.
    \item We further exploit temporal information by jointly recovering input actions and their timestamps, yielding a hierarchical spatiotemporal approach.
    \item Extensive experiments on simulation and real robotic manipulation demonstrate that our hierarchical spatiotemporal approach achieves superior performance over previous works.
\end{itemize}

\section{Related Work}
\label{sec:relatedwork}

\noindent \textbf{In-Context Imitation Learning.}
Robotic manipulation benefits from systems that can adapt to unseen tasks and scenes without retraining, making in-context imitation learning (ICIL) a promising paradigm for scalable robot learning~\cite{wang2025skil}. Rather than updating model parameters for new scenarios, ICIL allows robots to infer task structures directly from contextual demonstrations, inspired by the success of ICIL in large language models (LLMs)~\cite{brown2020language,mirchandani2023large,vosylius2023few,kwon2024language,vosylius2025instant}. Early approaches rely on state abstractions such as keypoint-based representations of objects and robots~\cite{di2024keypoint,papagiannis2024r,yin2024context}, enabling action inference from context but depending on external modules~\cite{caron2021emerging,xu2024flow} that generalize poorly across diverse settings. Recent works have shifted to end-to-end transformer frameworks~\cite{fu2024context,zhang2025dynamics}, where ICIL is modeled as a sequential prediction task and observation-action tokens are jointly learned. Building on this line of research, we introduce a hierarchical spatiotemporal action tokenizer (HiST-AT) to improve the performance and generalization of ICIL.

\noindent \textbf{Action Tokenization.}
Several research efforts have focused on learning robot action representations~\cite{chandak2019learning,zech2019action}. Initial methods~\cite{brohan2022rt,shafiullah2022behavior} discretize each action dimension into bins. ICRT~\cite{fu2024context} and CAPTURE~\cite{zhang2025dynamics} parameterize actions using neural networks, while Zhao et al.~\cite{zhao2023learning} learn low-dimensional action embeddings using a variational autoencoder (VAE). Vision–language–action (VLA) models represent robot actions as sequences of tokens~\cite{o2024open,huang2024emotion,kim2024openvla,pertsch2025fast}. However, these methods often lack smoothness, which is crucial for stable and successful robotic manipulation. Various approaches have been proposed to enhance smoothness, e.g., Lipschitz regularization~\cite{mysore2021regularizing}, Gaussian process priors~\cite{watson2023inferring}, action aggregation~\cite{zhao2023learning}, and Bayesian optimization~\cite{styrud2024bebop}. Recently, LipVQ-VAE~\cite{vuong2025action} introduces a vector-quantized VAE (VQ-VAE)~\cite{van2017neural}-based action tokenizer that relies on flat clustering and spatial reconstruction, enforcing smoothness via Lipschitz regularization. We propose a Lipschitz-smooth hierarchical spatiotemporal action tokenizer that leverages multi-level clustering and spatiotemporal reconstruction to learn smooth and effective action representations.

\noindent \textbf{Action Segmentation.}
Temporal action segmentation~\cite{kukleva2019unsupervised,xu2024temporally,spurio2025hierarchical} involves learning frame representations and clustering them into action segments. Early attempts~\cite{kukleva2019unsupervised,vidalmata2021joint} exploit temporal or spatiotemporal reconstruction for representation learning, while classical or hierarchical vector quantization has been applied for clustering in recent works~\cite{spurio2025hierarchical,gokay2025skeleton}. Moreover, PROGRESSOR~\cite{ayalew2025progressor} utilizes temporal reconstruction as a reward signal for learning robotic manipulation policy. Motivated by the aforementioned works, we develop a hierarchical spatiotemporal vector quantization framework for action tokenization.
\section{Our Approach}
\label{sec:method}

\subsection{In-Context Imitation Learning}
\begin{figure*}[t]
	\centering
		\includegraphics[width=0.85\linewidth, trim = 0mm 35mm 0mm 0mm, clip]{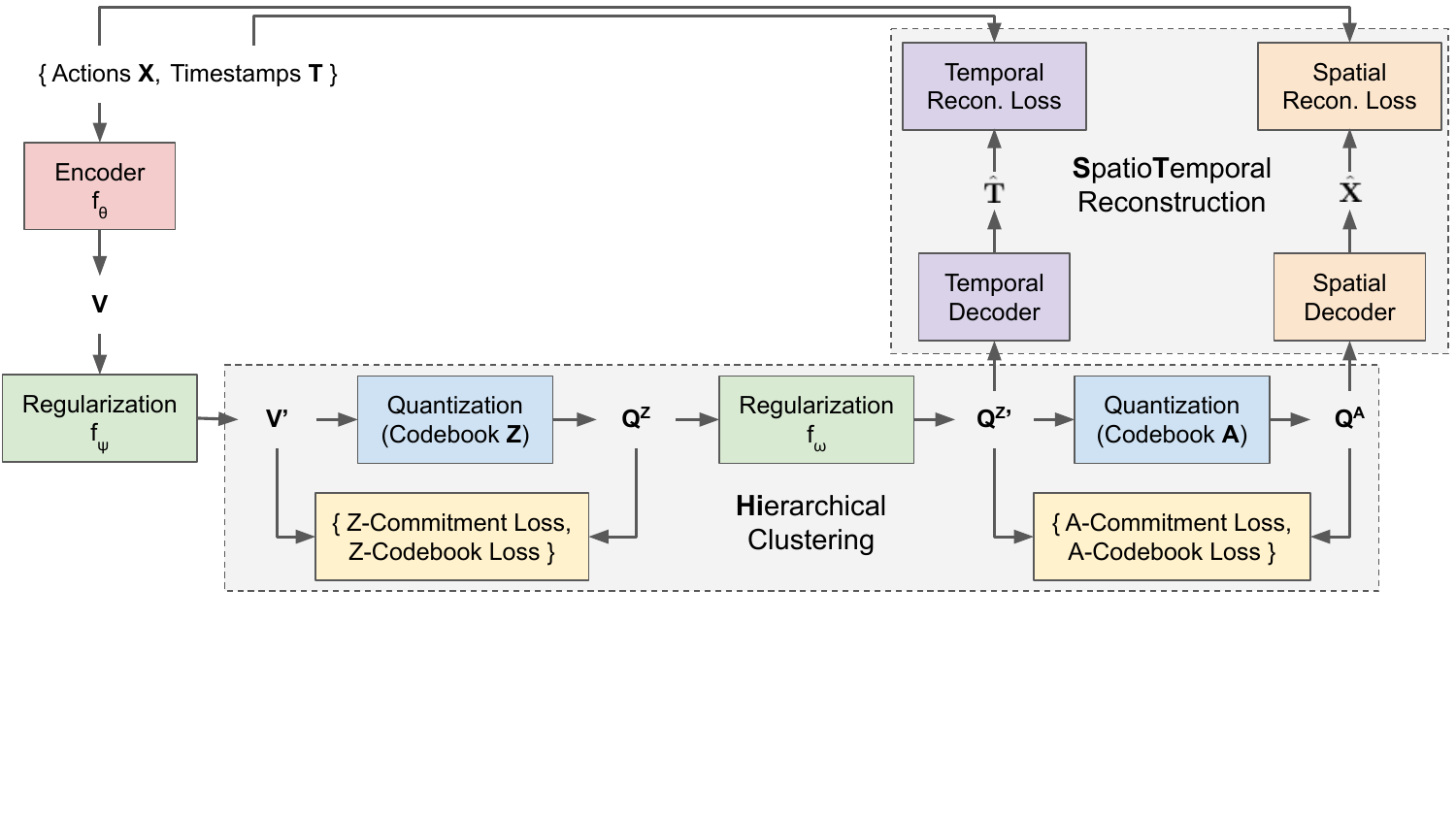}
	\caption{An overview of our hierarchical spatiotemporal action tokenizer (HiST-AT).}
	\label{fig:method}
\end{figure*}

ICIL aims to enable a policy to infer task behaviors from a small set of expert demonstrations provided at inference time without updating model weights. Following ICRT~\cite{fu2024context}, we model ICIL as a next-token prediction problem for robotic manipulation tasks (see Fig.~\ref{fig:teaser}(a)). We divide the expert demonstrations into two components: prompt demonstrations and query. Observations and actions are tokenized within the prompt demonstrations, yielding context rich prompt tokens for the model to execute robotic manipulation tasks. A transformer then auto-regressively processes these prompt tokens to predict a sequence of robot actions given query observations. The model can learn to adapt to unseen tasks by conditioning on prompt demonstrations without explicit retraining. Following~\cite{vuong2025action}, we employ ResNet-18~\cite{he2016deep} to encode RGB-D observations. Optionally, CLIP~\cite{radford2021learning} is used to encode language inputs, while MLPs are used to encode the rest of sensory inputs. An MLP maps the tokenized observations and actions into a shared latent space for dimensional consistency. We then utilize an autoregressive transformer that attends to a sequence of prompt observation and action tokens $(\mathbf{h}^\text{o}_{\text{p}},\mathbf{h}^\text{a}_{\text{p}})$ and query observation tokens $(\mathbf{h}^\text{o}_{\text{q}})$:
\begin{equation}
    \label{eq:icrt-sequence}
    \underbrace{(\mathbf{h}_{p1}^\text{o}, \mathbf{h}_{p1}^\text{a}, \dots, \mathbf{h}_{pM}^\text{o}, \mathbf{h}_{pM}^\text{a})}_{\text{prompt}},
    \underbrace{(\mathbf{h}_{q1}^\text{o}, \mathbf{h}_{q2}^\text{o}, \dots, \mathbf{h}_{qN}^\text{o})}_{\text{query}},
\end{equation}
where $M$ denotes the number of timestamps in the prompt demonstrations and $N$ represents the query timestamps. We use a full demonstration of a robotic task as prompt input as in~\cite{fu2024context}. Following the approach in~\cite{nasiriany2024robocasa}, the transformer is then trained by supervising predicted robot actions with the ground truth. The prompt tokens are masked while the unmasked query tokens are decoded via an MLP to generate robot actions. At inference time, the model autoregressively predicts the action by processing only one query observation at a time. ICRT~\cite{fu2024context} employs a simple MLP-based action tokenizer. However, this approach lacks smoothness. LipVQ-VAE~\cite{vuong2025action} employs Lipschitz regularization to enhance smoothness. In this work, we propose HiST-AT --- a \textbf{Hi}erarchical \textbf{S}patio\textbf{T}emporal \textbf{A}ction \textbf{T}okenizer. Our approach consists of two key modules: i) hierarchical clustering and ii) spatiotemporal reconstruction. Fig.~\ref{fig:method} illustrates an overview of HiST-AT.

\subsection{Hierarchical Spatiotemporal Action Tokenizer} 
\label{sec:model_details}
\noindent \textbf{Latent Representations of Robot Actions.}
We first employ an encoder $f_\theta$ that maps each input action $\mathbf{x}$ from a sequence of actions $\mathbf{X} \in \mathbb{R}^{(B \cdot S) \times D_{\text{feature}}}$ to a latent representation $\mathbf{v}$, producing a sequence of latent representations $\mathbf{V} \in \mathbb{R}^{(B \cdot S) \times D_{\text{hidden}}}$. Here, $B$ is the batch size, $S$ is the sequence length, $D_{\text{feature}}$ is the input dimension, and $D_{\text{hidden}}$ is the output dimension by the encoder. Following~\cite{tao2024maniskill3,nasiriany2024robocasa}, $\mathbf{x}$ consists of a robot action, including the relative position and angle of gripper. Inspired by~\cite{vuong2025action}, we use a Lipschitz-conditioned network $f_\psi$ to enforce smoothness in the latent representations $\mathbf{V}$, yielding the Lipschitz-regularized latent representations $\mathbf{V}^{\prime} \in \mathbb{R}^{(B \cdot S) \times D_{\text{latent}}}$, where $D_{\text{latent}}$ is the latent dimension. Specifically, each layer $\ell$ with weight $\mathbf{W}^{(\ell)}$ in the network $f_\psi$ is augmented with a trainable Lipschitz bound $c_\ell$, introduced for every row $i$ as follows:
\begin{equation}
\mathbf{W}_i^{(\ell)} = \frac{\mathbf{W}_i^{(\ell)}}{\sum_j |\mathbf{W}_{i,j}^{(\ell)}|} \cdot \operatorname{softplus}(c_\ell).
\end{equation}
where $\operatorname{softplus}(c_\ell) = \ln(1 + e^{c_\ell})$ enforces positivity of the Lipschitz bounds via reparameterization.

\noindent \textbf{Hierarchical Clustering.}
Inspired by HVQ~\cite{spurio2025hierarchical}, we present a hierarchical vector quantization framework to encode robot actions. Our vector quantization hierarchy consists of two learned codebooks 
$\mathbf{Z} = \{\mathbf{z}_j\}^{\alpha K}_{j=1}$ and 
$\mathbf{A} = \{\mathbf{a}_i\}^{K}_{i=1}$ corresponding to two levels of vector quantization. 
Here, $\mathbf{z}_j \in \mathbb{R}^{D_{latent}}$, $\mathbf{a}_i \in \mathbb{R}^{D_{latent}}$, $K$ is the number of codebook entries, and $\alpha$ is a ratio parameter. 
$\mathbf{A}$ represents $K$ action prototypes/clusters, while $\mathbf{Z}$ models $\alpha K$ subaction prototypes/clusters. 

The first vector quantization level maps each Lipschitz-regularized latent vector 
$\mathbf{v}_k^{\prime} \in \mathbf{V^\prime}$ to the closest prototype 
$\mathbf{z}_{j^*} \in \mathbf{Z}$ using $L_2$ distance, yielding the quantized $\mathbf{q}^Z_k$ as:
\begin{align}
    \mathbf{q}^Z_k = \mathbf{z}_{j^*},~~~\text{with}~~~j^* = \argmin_j ||\mathbf{v}_k^{\prime} - \mathbf{z}_j||_2.
\end{align}
Merging $\mathbf{q}^Z_k$ from all $\mathbf{v}_k^{\prime} \in \mathbf{V^{\prime}}$ yields the quantized 
$\mathbf{Q}^Z \in \mathbb{R}^{(B \cdot S) \times D_{latent}}$. 
We then pass $\mathbf{Q}^Z$ to a Lipschitz-conditioned network $f_\omega$, yielding the Lipschitz-smooth 
$\mathbf{Q}^{Z^\prime} \in \mathbb{R}^{(B \cdot S) \times D_{latent}}$. 
Similarly, the second vector quantization level maps each Lipschitz-regularized prototype 
$\mathbf{q}^{Z^{\prime}}_k \in \mathbf{Z}$ to the nearest prototype 
$\mathbf{a}_{i^*} \in \mathbf{A}$, yielding the quantized $\mathbf{q}^A_k$ as:
\begin{align}
    \mathbf{q}^A_k = \mathbf{a}_{i^*},~~~\text{with}~~~i^* = \argmin_i ||\mathbf{q}^{Z^{\prime}}_k - \mathbf{a}_i||_2.
\end{align}
Combining $\mathbf{q}^A_k$ from all $\mathbf{v}_k^{\prime} \in \mathbf{V}^{\prime}$ yields the quantized 
$\mathbf{Q}^A \in \mathbb{R}^{(B \cdot S) \times D_{latent}}$. Subaction quantization helps account for subtle motion variations, yielding more accurate action representations at the coarser level.
As discussed in Sec.~\ref{sec:experiments}, our hierarchical approach achieves superior performance over the non-hierarchical baseline~\cite{vuong2025action}. 

\noindent \textbf{Spatiotemporal Reconstruction.}
We propose spatiotemporal reconstruction, which exploits both spatial and temporal cues by jointly recovering input robot actions and associated timestamps,  inspired by CTE~\cite{kukleva2019unsupervised} and PROGRESSOR~\cite{ayalew2025progressor}. First for spatial reconstruction, we pass the quantized $\mathbf{Q}^A$ to the spatial decoder, which mirrors the encoder's architecture, producing the reconstructed robot actions $\mathbf{\hat{X}} \in \mathbb{R}^{(B \cdot S) \times D_{feature}}$. For temporal reconstruction, we pass the Lipschitz-regularized $\mathbf{Q}^{Z^{\prime}}$ to a temporal decoder with a simple architecture (i.e., an MLP network with two hidden layers), yielding the predicted timestamps $\mathbf{\hat{T}} \in \mathbb{R}^{B \cdot S}$. As shown in Sec.~\ref{sec:experiments}, our model outperforms the spatial reconstruction baseline~\cite{vuong2025action} by leveraging both spatial and temporal cues.

\subsection{Training Losses}
We train our model, including encoder, regularizers, subaction and action codebooks, and spatial and temporal decoders, by using a combination of hierarchical clustering, spatiotemporal reconstruction, and Lipschitz regularization losses. The codebooks are randomly initialized.

\noindent \textbf{Hierarchical Clustering.}
We use two commitment losses, corresponding to two quantization levels:
\begin{equation}
\begin{split}
\mathcal{L}_{\mathrm{commit}_Z} &= \frac{1}{B \cdot S} \sum_{k=1}^{B \cdot S} \left\| \mathbf{v}_k^{\prime} - \operatorname{sg}(\mathbf{q}_k^{Z}) \right\|_2^2, \\
\mathcal{L}_{\mathrm{commit}_A} &= \frac{1}{B \cdot S} \sum_{k=1}^{B \cdot S} \left\| \mathbf{q}^{Z^{\prime}}_k - \operatorname{sg}(\mathbf{q}^{A}_k) \right\|_2^2.
\end{split}
\end{equation}
Here, $\mathcal{L}_{\mathrm{commit}_Z}$ encourages the Lipschitz-regularized latent vector $\mathbf{v}_k^{\prime}$ to stay close to the assigned prototype $\mathbf{q}^Z_k$, while $\mathcal{L}_{\mathrm{commit}_A}$ pushes the Lipschitz-regularized prototype $\mathbf{q}^{Z^{\prime}}_k$ towards the chosen prototype $\mathbf{q}^A_k$. $\operatorname{sg}[\cdot]$ denotes the stop-gradient operator, and $B \cdot S$ is the total number of samples in $\mathbf{X}$. Next, we employ two codebook losses corresponding to the two quantization levels as:
\begin{equation}
\begin{split}
\mathcal{L}_{\mathrm{codebook}_Z} &= \frac{1}{B \cdot S} \sum_{k=1}^{B \cdot S} \left\| \operatorname{sg}(\mathbf{v}_k^{\prime}) - \mathbf{q}_k^{Z} \right\|_2^2, \\
\mathcal{L}_{\mathrm{codebook}_A} &= \frac{1}{B \cdot S} \sum_{k=1}^{B \cdot S} \left\| \operatorname{sg}(\mathbf{q}^{Z^{\prime}}_k) - \mathbf{q}^{A}_k \right\|_2^2.
\end{split}
\end{equation}
Here, $\mathcal{L}_{\mathrm{codebook}_Z}$ encourages the assigned prototype $\mathbf{q}_k^Z$ to stay close to the Lipschitz-regularized latent vector $\mathbf{v}^{\prime}_k$, while $\mathcal{L}_{\mathrm{codebook}_A}$ pushes the chosen prototype $\mathbf{q}_k^A$ towards the corresponding Lipschitz-regularized prototype $\mathbf{q}_k^{Z^{\prime}}$. Lastly, subcluster and cluster level losses are written as:
\begin{equation}
\begin{split}
\mathcal{L}_{\mathrm{vq}_Z} &= \mathcal{L}_{\mathrm{commit}_Z} + \mathcal{L}_{\mathrm{codebook}_Z}, \\
\mathcal{L}_{\mathrm{vq}_A} &= \mathcal{L}_{\mathrm{commit}_A} + \mathcal{L}_{\mathrm{codebook}_A}.
\end{split}
\end{equation}

\noindent \textbf{Spatiotemporal Reconstruction.}
We measure the spatial reconstruction loss between reconstructed actions $\hat{\mathbf{X}}$ and original actions $\mathbf{X}$ and the temporal reconstruction loss between predicted timestamps $\hat{\mathbf{T}}$ and original timestamps $\mathbf{T}$ by adopting Mean Squared Error (MSE), defined as:
\begin{equation}
\begin{split}
\mathcal{L}_{\text{spat}} &= \frac{1}{B \cdot S} \sum_{k=1}^{B \cdot S} \left\| \hat{\mathbf{X}}^{(k)} - \mathbf{X}^{(k)} \right\|_2^2, \\
\mathcal{L}_{\text{temp}} &= \frac{1}{B \cdot S} \sum_{k=1}^{B \cdot S} \left\| \hat{\mathbf{T}}^{(k)} - \mathbf{T}^{(k)} \right\|_2^2.
\end{split}
\end{equation}

\noindent \textbf{Final Loss.} Our final loss combines the above losses and the Lipschitz regularization losses~\cite{vuong2025action}: 
\begin{equation}
\begin{split}
\mathcal{L} = {}& \lambda_{vq}(\mathcal{L}_{\mathrm{vq}_Z}+ \mathcal{L}_{\mathrm{vq}_A}) + \lambda_{spat}\mathcal{L}_{\mathrm{spat}} \\
& + \lambda_{temp}\mathcal{L}_{\mathrm{temp}} + \lambda_{reg}(\mathcal{L}_{\mathrm{reg}_Z}+ \mathcal{L}_{\mathrm{reg}_A}).
\end{split}
\end{equation}
Here, $\lambda_{vq}$ is the weight for hierarchical clustering losses, $\lambda_{spat}$ and $\lambda_{temp}$ are the weights for spatiotemporal reconstruction losses, and $\lambda_{reg}$ is the weight for regularization terms. Using a more advanced temporal loss or decoder may further boost performance, which we leave for future work.
\section{Experiments}
\label{sec:experiments}

\subsection{Simulation Robotic Manipulation Results}

\noindent \textbf{Experiment Settings.}
We conduct simulation experiments in RoboCasa~\cite{nasiriany2024robocasa} and ManiSkill~\cite{tao2024maniskill3} on a single NVIDIA A100 GPU. In RoboCasa, seven tasks are evaluated with training for 500K iterations under the standard protocol, while in ManiSkill we focus on three tasks, training for 30K iterations to remain consistent with prior works. We evaluate the performance through success rate, as defined by each environment~\cite{nasiriany2024robocasa, tao2024maniskill3}. Our HiST-AT tokenizer builds on LipVQ-VAE~\cite{vuong2025action}, with a two-layer MLP encoder projected via a Lipschitz-regularized layer to $D_{latent}=208$, processing action chunks of sequence length $S=16$ for RoboCasa and $S=30$ for ManiSkill. Latents are hierarchically quantized via two lookup-free codebooks, $N_z=64$ sub-action and $N_a=16$ action prototypes, with a mirrored decoder reconstructing actions and a 3-layer sigmoid MLP temporal decoder predicting normalized timesteps from the quantized latents. Training combines MSE reconstruction loss, commitment/codebook losses (weight $0.25$), and a temporal loss (weight $0.02$). We compare our ICRT~\cite{fu2024context}-based framework with established approaches such as BC-Transformer~\cite{mandlekar2021matters}, ACT~\cite{zhao2023learning}, and MCR~\cite{jiang2025robots}; since ACT is computationally intensive and exceeds our hardware capacity, we scale it down to match BC-Transformer for fair comparisons. Within the same ICRT~\cite{fu2024context}-based framework, we also evaluate against other action tokenizers, including MLP~\cite{fu2024context}, discrete binning~\cite{brohan2022rt}, FAST~\cite{pertsch2025fast}, VQ-VAE~\cite{van2017neural}, LFQ-VAE~\cite{yu2024language}, and LipVQ-VAE~\cite{vuong2025action}, where FAST is further fine-tuned on one million action samples from RoboCasa and ManiSkill. 

\begin{table*}[t]
    \centering
    \caption{Robotic manipulation results on RoboCasa~\cite{nasiriany2024robocasa}.}\label{tab:robocasa}
    \setlength{\tabcolsep}{4pt} 
    \scriptsize
    \renewcommand{\arraystretch}{0.85}{
        \begin{tabular}{@{}r|ccccccc|c@{}}
            \toprule
            \diagbox{Method}{Task} & \shortstack{Pick \\ and \\ Place} & \shortstack{Open \\ Close \\ Doors} & \shortstack{Open \\ Close \\ Drawers} & \shortstack{Turning \\ Levers \\ $ $} & \shortstack{Twisting \\ Knobs \\ $ $} & \shortstack{Insertion \\ $ $ \\ $ $} & \shortstack{Pressing \\ Buttons \\ $ $} & \shortstack{Average \\ $ $ \\ $ $}\\
            \midrule
            MCR~\cite{jiang2025robots}  &  0.00 & 0.31 & 0.18 & 0.17 & 0.02 & 0.01 & 0.22 & 0.120  \\
            ACT~\cite{zhao2023learning} & 0.01 & 0.13 & 0.17 & 0.15 & 0.12 & 0.07 & 0.06 & 0.083 \\
            
            BC-Transformer~\cite{mandlekar2021matters} & 0.29 & 0.55 & 0.78 & 0.62 & 0.31 & 0.24 & \textbf{0.78} & 0.477\\
            \midrule
            ICRT~\cite{fu2024context}+MLP~\cite{fu2024context} & 0.20 & 0.61 & 0.81 & 0.70 & 0.32 & 0.35 & 0.64 & 0.442 \\
            ICRT+Bin~\cite{brohan2022rt} & 0.25 & 0.75 & 0.78 & \textbf{0.81} & 0.32 & 0.34 & 0.59 & 0.483\\
            ICRT+FAST~\cite{pertsch2025fast}  & 0.30 & 0.59 & 0.80 & 0.57 & 0.39 & 0.19 & 0.63 & 0.471\\
            ICRT+VQ-VAE~\cite{van2017neural} & 0.20 & 0.70 & 0.84 & 0.77 & 0.27 & 0.18 & 0.70 & 0.475\\
            ICRT+LFQ-VAE~\cite{yu2024language} & 0.27 & 0.69 & 0.83 & 0.77 & 0.40 & 0.27 & 0.68 & 0.489\\
            ICRT+LipVQ-VAE~\cite{vuong2025action} & 0.32 & 0.80 & 0.84 & 0.68 & 0.41 & 0.41 & 0.59 & 0.530\\
            \midrule
            \rowcolor{beaublue}ICRT+HiST-AT (Ours) & \textbf{0.35} & \textbf{0.90} & \textbf{0.89} & 0.72 & \textbf{0.52} & \textbf{0.44} & 0.63 & \textbf{0.590}\\
            \bottomrule
        \end{tabular}
    }
\end{table*}
\begin{table}[t]
    \centering
    \caption{Robotic manipulation results on ManiSkill~\cite{tao2024maniskill3}.}\label{tab:maniskill}
    \setlength{\tabcolsep}{5pt} 
    \scriptsize
    \renewcommand{\arraystretch}{0.85}{
        \resizebox{\columnwidth}{!}{
        \begin{tabular}{r|ccc|c}
            \toprule
            \diagbox{Method}{Task} & Pick Cube & Push Cube & Stack Cube & Average\\
            \midrule
            MCR~\cite{jiang2025robots} & 0.56 & 0.51 & 0.11 & 0.393 \\
            ACT~\cite{zhao2023learning} & 0.20 & 0.76 & 0.30 & 0.420 \\
            BC-Transformer~\cite{mandlekar2021matters} & 0.04 & \textbf{0.98} & 0.14 & 0.387 \\
            \midrule
            ACT~\cite{zhao2023learning}+Bin~\cite{brohan2022rt} & 0.71 & 0.52 & 0.25 & 0.493 \\
            ACT+FAST~\cite{pertsch2025fast} & 0.70 & 0.48 & 0.25 & 0.477 \\
            ACT+VQ-VAE~\cite{van2017neural} & 0.64 & 0.80 & 0.21 & 0.550 \\
            ACT+LFQ-VAE~\cite{yu2024language} & 0.74 & 0.70 &  0.23 & 0.557 \\
            ACT+LipVQ-VAE~\cite{vuong2025action} & 0.78 & 0.77 & 0.30 & 0.617 \\
            \midrule
            \rowcolor{beaublue}ACT+HiST-AT (Ours) & \textbf{0.85} & 0.78 & \textbf{0.38} & \textbf{0.670}\\
            \bottomrule
        \end{tabular}
        }
    }
\end{table}

\noindent \textbf{Results on RoboCasa.}
We evaluate on the MimicGen~\cite{mandlekar2023mimicgen} dataset in RoboCasa, with the results presented in Tab.~\ref{tab:robocasa}. The results show that our method significantly enhances performance, achieving average success rate of 59\% compared to 53\% of the previous best LipVQ-VAE~\cite{vuong2025action}. Moreover, incorporating hierarchical clustering and spatiotemporal reconstruction increases the overall effectiveness of our method, demonstrated by a 14.8\% performance gap between our method HiST-AT and the lowest performing MLP~\cite{fu2024context}. Overall, HiST-AT outperforms prior action tokenizers, including FAST~\cite{pertsch2025fast} which further has language inputs and is fine-tuned on one million action samples.

\noindent \textbf{Results on ManiSkill.}
To examine generalization beyond ICIL, we modify the cVAE-based encoder in ACT~\cite{zhao2023learning} with different action tokenizers, including our HiST-AT. Also, following LipVQ-VAE~\cite{vuong2025action}, we add a depth channel and train MCR~\cite{jiang2025robots} jointly with the policy head. The results in Tab.~\ref{tab:maniskill} show that HiST-AT achieves the best overall performance, surpassing the prior best LipVQ-VAE by 5.3\%. While previous approaches such as ACT and LipVQ-VAE attempt to address action smoothness, their limitations in modeling hierarchical structure and temporal consistency restrict their performance, whereas HiST-AT effectively captures both, leading to notable improvements.

\subsection{Ablation Results}

\begin{table*}[t]
    \centering
    \caption{Impacts of model components on RoboCasa~\cite{nasiriany2024robocasa}.}\label{tab:model_components}
    \setlength{\tabcolsep}{4pt} 
    \scriptsize
    \renewcommand{\arraystretch}{0.85}{
        \begin{tabular}{r|ccccccc|c}
            \toprule
            \diagbox{Components}{Task} & \shortstack{Pick \\ and \\ Place} & \shortstack{Open \\ Close \\ Doors} & \shortstack{Open \\ Close \\ Drawers} & \shortstack{Turning \\ Levers \\ $ $} & \shortstack{Twisting \\ Knobs \\ $ $} & \shortstack{Insertion \\ $ $ \\ $ $} & \shortstack{Pressing \\ Buttons \\ $ $} & \shortstack{Average \\ $ $ \\ $ $}\\
            \midrule
            Baseline~\cite{vuong2025action} & 0.32 & 0.80 & 0.84 & 0.68 & 0.41 & 0.41 & 0.59 & 0.530\\
            \midrule
            w/ Spatiotemporal Reconstruction & 0.33 & 0.82 & \textbf{0.90} & 0.68 & 0.42 & 0.42 & 0.61 & 0.552\\
            w/ Hierarchical Clustering & \textbf{0.36} & 0.85 & 0.86 & 0.70 & 0.47 & 0.42 & 0.62 & 0.573\\
          
            \rowcolor{beaublue}w/ Both & 0.35 & \textbf{0.90} & 0.89 & \textbf{0.72} & \textbf{0.52} & \textbf{0.44} & \textbf{0.63} & \textbf{0.590}\\
            \bottomrule
        \end{tabular}
    }
\end{table*}
\begin{figure}[t]
	\centering
		\includegraphics[width=0.85\linewidth]{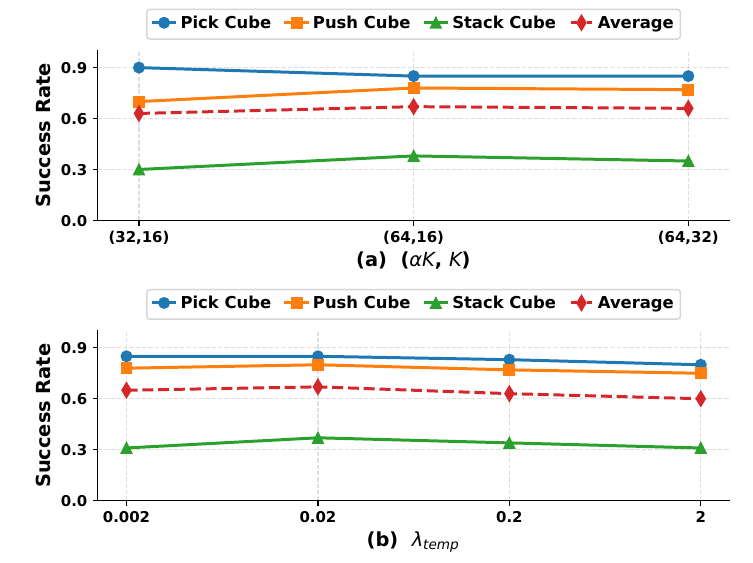}
	\caption{Impacts of (a) codebook sizes and (b) $\lambda_{temp}$ on ManiSkill~\cite{tao2024maniskill3}.}
	\label{fig:ablation_params}
\end{figure}
\begin{table*}[t]
    \centering
    \caption{Cross-dataset robotic manipulation results (MimicGen~\cite{mandlekar2023mimicgen}$\to$Human~\cite{nasiriany2024robocasa}).}\label{tab:cross-dataset}
    \setlength{\tabcolsep}{4pt} 
    \scriptsize
    \renewcommand{\arraystretch}{0.85}{
        \begin{tabular}{r|ccccccc|c}
            \toprule
            \diagbox{Method}{Task} & \shortstack{Pick \\ and \\ Place} & \shortstack{Open \\ Close \\  Doors} & \shortstack{Open \\ Close \\ Drawers} & \shortstack{Turning \\ Levers \\ $ $} & \shortstack{Twisting \\ Knobs \\ $ $} & \shortstack{Insertion \\ $ $ \\ $ $} & \shortstack{Pressing \\ Buttons \\ $ $}  & \shortstack{Average \\ $ $ \\ $ $} \\
            \midrule
            MCR~\cite{jiang2025robots} & 0.00 & 0.24 & 0.20 & 0.12 & 0.00 & 0.00 & 0.19 & 0.096 \\
            ACT~\cite{zhao2023learning} & 0.00 & 0.10 & 0.15 & 0.13 & 0.09 & 0.08 & 0.05 & 0.066 \\
            BC-Transformer~\cite{mandlekar2021matters} & 0.25 & 0.41 & 0.73 & 0.62 & 0.28 & 0.18 & 0.64 & 0.408 \\
            \midrule
            ICRT~\cite{fu2024context}+MLP~\cite{fu2024context} & 0.21 & 0.61 & \textbf{0.87} & 0.77 & 0.30 & 0.36 & 0.52 & 0.457 \\
            ICRT+Bin~\cite{brohan2022rt} & 0.26 & 0.75 & 0.79 & 0.74 & 0.31 & 0.29 & 0.60 & 0.495 \\
            ICRT+FAST~\cite{pertsch2025fast} & 0.30  & 0.63 & 0.77 & 0.74 & 0.36 & 0.39 & 0.42 & 0.481 \\
            ICRT+VQ-VAE~\cite{van2017neural} & 0.21 & 0.70 & 0.83 & 0.77 & 0.36 & 0.31 & 0.60 & 0.483 \\
            ICRT+LFQ-VAE~\cite{yu2024language} & 0.29  & 0.65 & 0.82 & 0.79 & 0.38 & 0.28 & 0.62 & 0.503 \\
            ICRT+LipVQ-VAE~\cite{vuong2025action} & 0.32 & 0.76 & 0.83 & 0.71 & 0.38 & 0.42 & 0.64 & 0.525 \\
            \midrule
            \rowcolor{beaublue}ICRT+HiST-AT (Ours) & \textbf{0.40} & \textbf{0.79} & 0.73 & \textbf{0.80} & \textbf{0.40} & \textbf{0.48} & \textbf{0.67} & \textbf{0.582} \\
            \bottomrule
        \end{tabular}
    }
\end{table*}
\begin{table*}[t]
    \centering
    \caption{Zero-shot robotic manipulation results on RoboCasa~\cite{nasiriany2024robocasa}.}\label{tab:zero-shot-robocasa}
    \setlength{\tabcolsep}{4pt} 
    \scriptsize
    \renewcommand{\arraystretch}{0.85}
    {
        \begin{tabular}{r|ccccccc|c}
            \toprule
            \diagbox{Method}{Task} & \shortstack{Pick \\ and \\ Place} & \shortstack{Open \\ Close \\  Doors} & \shortstack{Open \\ Close \\ Drawers} & \shortstack{Turning \\ Levers \\ $ $} & \shortstack{Twisting \\ Knobs \\ $ $} & \shortstack{Insertion \\ $ $ \\ $ $} & \shortstack{Pressing \\ Buttons \\ $ $}  & \shortstack{Average \\ $ $ \\ $ $} \\
            \midrule
            MCR~\cite{jiang2025robots} & 0.00 & 0.00 & 0.00  & 0.06 & 0.04 & 0.01 & 0.00 & 0.021  \\
            ACT~\cite{zhao2023learning} & 0.00 & 0.00 & 0.00  & 0.05 & 0.04 & 0.01 & 0.00 & 0.019  \\
            BC-Transformer~\cite{mandlekar2021matters} & 0.00 & 0.00 & 0.00 & 0.10 & 0.20 & 0.00 & 0.00 & 0.028 \\
            \midrule
            ICRT~\cite{fu2024context}+MLP~\cite{fu2024context} & 0.00 & 0.01 & 0.00 & 0.19 & 0.22 & 0.01 & 0.02 & 0.047 \\
            ICRT+Bin~\cite{brohan2022rt} & 0.01 & 0.01 & 0.00 & 0.16 & 0.24 & 0.01 & 0.02 & 0.046 \\
            ICRT+FAST~\cite{pertsch2025fast} & 0.00 & 0.00 & 0.00 & 0.20 & 0.12 & 0.01 & 0.00 & 0.041 \\
            ICRT+VQ-VAE~\cite{van2017neural} & 0.00 & 0.00 & 0.00 & 0.16 & 0.20 & 0.01 & 0.02 & 0.041 \\
            ICRT+LFQ-VAE~\cite{yu2024language} & 0.01 & 0.01 & 0.00 & 0.24 & 0.14 & 0.02 & 0.02 & 0.049 \\
            ICRT+LipVQ-VAE~\cite{vuong2025action} & 0.01 & 0.00 & 0.00 & 0.25 & 0.24 & 0.00 & 0.02 & 0.052 \\
            \midrule
            \rowcolor{beaublue}ICRT+HiST-AT (Ours) & \textbf{0.01} & \textbf{0.01} & \textbf{0.00} & \textbf{0.26} & \textbf{0.28} & \textbf{0.03} & \textbf{0.21} & \textbf{0.090} \\
            \bottomrule
        \end{tabular}
    }
\end{table*}

\noindent \textbf{Impacts of Model Components.}
We analyze the contribution of each component in our method on Robocasa~\cite{nasiriany2024robocasa} in Tab.~\ref{tab:model_components}. Starting from the baseline LipVQ-VAE~\cite{vuong2025action}, adding hierarchical clustering improves success rates significantly, highlighting the benefit of modeling structured action hierarchies, while integrating spatiotemporal reconstruction instead yields smaller gains. Incorporating both components in our HiST-AT performs the best, achieving 6\% average performance increase compared to the baseline. These results demonstrate that hierarchical clustering and spatiotemporal reconstruction provide complementary gains, yielding superior performance over the baseline.

\noindent \textbf{Impacts of Codebook Sizes.}
We investigate the effect of the sizes of the codebooks $\mathbf{Z}$ and $\mathbf{A}$, i.e., ($\alpha K$,$K$) respectively, on ManiSkill~\cite{tao2024maniskill3}. As shown in Fig.~\ref{fig:ablation_params}(a), increasing from (32,16) to (64,16) improves overall performance, indicating that a larger number of subaction clusters helps capture fine-grained action dynamics. However, further increasing to (64,32) does not provide additional gains, suggesting redundancy in representation. Overall, (64,16) offers the best tradeoff between capturing high-level structures and detailed action variations, and is used in all of our experiments.

\noindent \textbf{Impacts of $\lambda_{temp}$.}
We analyze the effect of the temporal reconstruction weight $\lambda_{temp}$ on ManiSkill~\cite{tao2024maniskill3} by varying its value in the range [0.002, 2], as shown in Fig.~\ref{fig:ablation_params}(b). The results indicate that moderate temporal supervision is most effective, i.e., $\lambda_{temp}$ = 0.02 achieves the strongest overall performance. Larger weights lead to a decline in performance, suggesting that excessive emphasis on timestamp prediction can hinder the learning of action representations. Overall, the results show that a balanced temporal reconstruction weight is crucial for capturing action dynamics without overwhelming the primary learning objective, and we set $\lambda_{temp}$ = 0.02 for all of our experiments.

\noindent \textbf{Cross-Dataset Results.}
We evaluate transfer from the MimicGen dataset~\cite{mandlekar2023mimicgen} to the Human dataset~\cite{nasiriany2024robocasa} in RoboCasa, containing sparser and less structured object arrangements. The results in Tab.~\ref{tab:cross-dataset} show that ICRT~\cite{fu2024context}-based methods demonstrate stronger robustness as compared to BC-Transformer~\cite{mandlekar2021matters}. Even with an MLP action tokenizer~\cite{fu2024context}, the ICRT framework surpasses BC-Transformer. More importantly, our HiST-AT further improves cross-dataset performance, outperforming the second best LipVQ-VAE~\cite{vuong2025action} by 6\% on average, highlighting the benefit of hierarchical clustering and spatiotemporal reconstruction in capturing transferable action representations.

\noindent \textbf{Zero-Shot Results.}
To further evaluate generalization to unseen data, we perform zero-shot experiments by training on a subset of tasks and testing on another, following the split in RoboCasa~\cite{nasiriany2024robocasa}. The results in Tab.~\ref{tab:zero-shot-robocasa} show that ICRT~\cite{fu2024context}-based methods outperform other approaches like BC-Transformer~\cite{mandlekar2021matters}. Moreover, our HiST-AT performs the best, surpassing the second best LipVQ-VAE~\cite{vuong2025action} by 3.8\% on average, demonstrating stronger generalization to unseen action sequences.

\noindent \textbf{Impacts of Input to Temporal Decoder.}
To evaluate the impact of different input representations on temporal reasoning, we feed three variants of latent prototypes $\mathbf{Q}^Z$, $\mathbf{Q}^{Z^\prime}$, and $\mathbf{Q}^A$ into the temporal MLP decoder, where $\mathbf{Q}^Z$ is output of the first quantization codebook Z, $\mathbf{Q}^{Z^\prime}$ is regularized output derived from $\mathbf{Q}^Z$ before hierarchical clustering and $\mathbf{Q}^A$ is output of the second quantization codebook A after hierarchical clustering. Tab.~\ref{table:temporal_decoder} reports the success rates for each input on three ManiSkill ~\cite{tao2024maniskill3} tasks: Pick Cube, Push Cube, and Stack Cube. $\mathbf{Q}^{Z^\prime}$ achieves 68\%, 76\% and 28\% success rates as compared to $\mathbf{Q}^{Z}$ (37\%, 74\% and 23\%) and $\mathbf{Q}^{A}$ (54\%, 74\% and 24\%) for Pick Cube, Stack Cube and Push Cube tasks respectively. The performance gain from using $\mathbf{Q}^{Z^\prime}$ can be attributed to  the regularization applied to $\mathbf{Q}^{Z}$ which reduces temporal jitter and enforces smoother latent representations, enhancing the decoder's ability to model long range dependencies, while preserving the low-level action cues in $\mathbf{Q}^{Z}$ without its quantization nose.
\begin{table}[t]
    \centering
    \scriptsize
    \caption{Impacts of input to temporal decoder.}
    \label{table:temporal_decoder}
    \vspace{1ex}
    \setlength{\tabcolsep}{6pt} 
    \renewcommand{\arraystretch}{1.2} 
    \begin{tabular}{l|ccc}
        \toprule
        Input & Pick Cube & Push Cube & Stack Cube \\
        \midrule
        $\mathbf{Q}^Z$     & 37\% & 74\% & 23\% \\
        \rowcolor{beaublue}$\mathbf{Q}^{Z^\prime}$    & \textbf{68\%} & \textbf{76\%} & \textbf{28\%} \\
        $\mathbf{Q}^A$     & 54\% & 72\% & 24\% \\
        \bottomrule
    \end{tabular}
\end{table}

\noindent \textbf{Impacts of Input to Spatial Decoder.} We analyze the effect of spatial decoder placement in the model architecture in Fig.~\ref{fig:method} by passing different vector quantization outputs (i.e $\mathbf{Q}^{Z^\prime}$, $\mathbf{Q}^{A}$, or both) to the spatial decoder and include the results across three tasks of Maniskill~\cite{tao2024maniskill3} in Tab.~\ref{table:spatial_decoder}. From the results, we observe that the spatial decoder performs best when the output of the second vector quantization $\mathbf{Q}^A$ is used as input. We see a slight decrease in performance when both $\mathbf{Q}^{Z^\prime}$ and $\mathbf{Q}^A$ are used. A sharp decline in performance is noted when only $\mathbf{Q}^{Z^\prime}$ is used as the input.  
\begin{table}[t]
    \centering
    \scriptsize
    \caption{Impacts of input to spatial decoder.}
    \label{table:spatial_decoder}
    \vspace{1ex}
    \setlength{\tabcolsep}{6pt} 
    \renewcommand{\arraystretch}{1.2} 
    \begin{tabular}{c|ccc}
        \toprule
        Input & Pick Cube & Push Cube & Stack Cube \\
        \midrule
        $\mathbf{Q}^{Z^\prime}$ & 38\% & 70\% & 21\% \\
        \rowcolor{beaublue}$\mathbf{Q}^{A}$    & \textbf{68\%} & \textbf{76\%} & \textbf{28\%} \\
        Both & 65\% & 74\% & 25\% \\
        \bottomrule
    \end{tabular}
\end{table}

\noindent \textbf{Smoothness Comparisons.}
To evaluate the smoothness of our proposed HiST-AT framework, we extract 500 action latent representation from the RoboCasa~\cite{nasiriany2024robocasa} benchmark and compute the smoothness score using the least-energy function formulation introduced in prior work ~\cite{horn1983curve}. This metric measures trajectory by estimating the energy required to traverse latent transitions, where lower energy indicates more continuous and stable temporal representations. Tab.~\ref{table:smoothness_eval} summarizes the smoothness performance across different action tokenization approaches. Our method, HiST-AT, achieves a smoothness score of 0.11, demonstrating its ability to maintain coherent temporal transitions in the latent action space. This suggests that HiST-AT captures structured action progression while preserving temporal consistency. Overall, smooth latent action representation remains an important component of effective manipulation learning.

\begin{table}[t]
    \centering
    \caption{Smoothness comparisons.}
    \label{table:smoothness_eval}
    \vspace{1ex}
    \setlength{\tabcolsep}{3pt} 
    \resizebox{\linewidth}{!}{
        \begin{tabular}{l|cccccc>{\columncolor{beaublue}}c}
            \toprule
            Tokenizer & \shortstack{ MLP \\ ~\cite{fu2024context}} & \shortstack{Bin\\ ~\cite{brohan2022rt}} & \shortstack{FAST\\ ~\cite{pertsch2025fast}} & \shortstack{VQ-VAE \\ ~\cite{van2017neural}} & \shortstack{LFQ-VAE\\ ~\cite{yu2024language}}  & \shortstack{LipVQ-VAE \\ ~\cite{vuong2025action}} & \shortstack{HiST-AT \\ (Ours)} \\
            \hline
            Smoothness $\downarrow$ & 5.71 & 12.57 & 5.67 & 4.79 & 2.34 & 0.63 & \textbf{0.11} \\
            \bottomrule
        \end{tabular}
    }
\end{table}

\noindent \textbf{Run Time Comparisons.}
All experiments were conducted on NVIDIA A100 GPUs for both training and evaluation on the RoboCasa~\cite{nasiriany2024robocasa}. As summarized in Tab.~\ref{tab:runtime_comparison}, our approach requires 5.3 minutes per epoch. In comparison, LipVQ-VAE requires 4.5 minutes per epoch. However, LipVQ-VAE often converges at around 900 epochs, yielding roughly 67 hours of training, whereas our approach usually converges at around 500 epochs, yielding roughly 44 hours of training. For testing, both methods require approximately 37 minutes per RoboCasa subtask. These results demonstrate better training efficiency, making it a more computationally practical solution for large scale robot manipulation learning.

\begin{table}[t]
\centering
\caption{Runtime comparison between methods.}
\label{tab:runtime_comparison}
\begin{tabular}{l|c|c}
\hline
Method & \makecell{Training Time \\ per epoch} & \makecell{Evaluation Time \\ per sub-task}\\
\hline
LipVQ-VAE~\cite{vuong2025action} &  4.5 min & 37 min \\
\rowcolor{beaublue}HiST-AT (Ours) & 5.3 min & 37 min \\
\hline
\end{tabular}
\end{table}

\subsection{Real-World Robotic Manipulation Results}

\begin{figure}[!ht]
\centering
\includegraphics[width=0.85\columnwidth, trim = 0mm 85mm 115mm 0mm, clip]{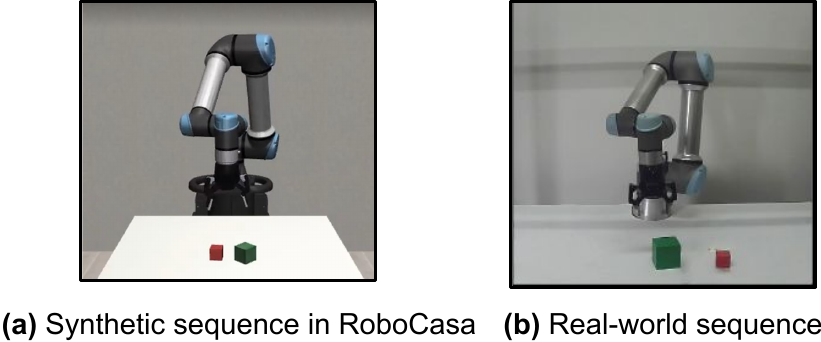}
\caption{Sim-to-real examples.}
\label{fig:sim2real}

\vspace{1ex}

\setlength{\tabcolsep}{5pt} 
\scriptsize
\renewcommand{\arraystretch}{0.85}{
\begin{tabular}{lcc}
    \toprule
    \diagbox{Method}{Task} &  \shortstack{Pick \\ Cube} &  \shortstack{Stack \\ Cube}
    \\ \midrule
    ICRT+LipVQ-VAE~\cite{vuong2025action} & 0.19 & 0.12 \\ \midrule
    \rowcolor{beaublue}ICRT+HiST-AT (Ours) & \textbf{0.23} & \textbf{0.14} \\
    \bottomrule
\end{tabular}
}
\captionof{table}{Sim-to-real results.}
\label{tab:sim2real}
\end{figure}

We gather demonstrations in RoboCasa using a UR5e arm for Pick Cube and Stack Cube. For each task, we collect 10 teleoperation demonstrations. We then use MimicGen to generate 2,000 synthetic demonstrations per task. Separately, we collect 10 real demonstrations per task and combine synthetic and real demonstrations to form a training dataset (see examples in Fig.~\ref{fig:sim2real}). Data is recorded from a third-person view and an in-hand view, with the simulation controller synchronized to the real UR5e controller. For evaluation, we collect 50 demonstrations per task. Tab.~\ref{tab:sim2real} presents the results. Despite the expanded dataset, sim-to-real transfer remains challenging; however, our method produces smoother motions and achieves higher success rates than LipVQ-VAE~\cite{vuong2025action}.

\section{Limitations}
\label{sec:limitations}
Our experiments on RoboCasa ~\cite{nasiriany2024robocasa} and ManiSkill ~\cite{tao2024maniskill3} demonstrate that HiST-AT significantly outperforms prior action tokenizers. Ablation studies further reveal that both hierarchical clustering and spatiotemporal reconstruction provide complementary gains, while real-world trials confirm successful transfer to physical manipulation tasks. However, our approach sees some limitations. The fixed action sequence length $S$ constrains action representations to a pre-determined granularity, limiting adaptability to the variable temporal extent of sub-action primitives across tasks. This rigidity likely also contributes to our modest zero-shot performance, and adaptive temporal segmentation is a promising direction for future work. The two-level hierarchy adds per-step overhead that scales with codebook size, potentially limiting large-vocabulary or low-latency applications despite faster overall convergence. Performance on complex tasks such as Stack Cube remain limited, reflecting persistent sim-to-real gaps. Finally, like other action tokenizers, HiST-AT does not resolve the broader challenge of fusing hierarchical action representations with multimodal observations such as vision, language, and proprioception. Our future work will explore more advanced temporal objectives. 
\section{Conclusion}
\label{sec:conclusion}

We propose an in-context imitation learning framework based on HiST-AT, a hierarchical spatiotemporal action tokenizer. Specifically, we introduce a two-level vector quantization hierarchy, where input actions are mapped to fine-grained subclusters at the lower level and further grouped into higher-level clusters. Our hierarchical approach outperforms the non-hierarchical baseline while primarily relying on spatial information via action reconstruction. We further incorporate spatial and temporal cues by jointly reconstructing actions and timestamps within the multi-level hierarchy, yielding a hierarchical spatiotemporal action tokenizer. Extensive evaluations on simulation and real-world robotic manipulation benchmarks demonstrate superior performance over prior methods. Our future work will explore more advanced temporal objectives to further improve performance.

{\small
\bibliographystyle{IEEEtran}
\bibliography{references}
}

\end{document}